\documentclass[10pt,twocolumn,letterpaper]{article}

\usepackage{cvpr}
\usepackage{epsfig}
\usepackage{graphicx}
\usepackage{amsmath}
\usepackage{amssymb}

\usepackage{multirow}
\usepackage{xcolor, url}

\cvprfinalcopy

\def\cvprPaperID{} %

\ifcvprfinal\fi
\begin{document}

\title{Real-time Semantic Image Segmentation via Spatial Sparsity}

\author{Zifeng Wu, Chunhua Shen\thanks{Corresponding author.}, and Anton van den Hengel\\
The University of Adelaide,
Adelaide, SA 5005, Australia\\
{\tt\small firstname.lastname@adelaide.edu.au}
}

\maketitle
\begin{abstract}
We propose an approach to semantic (image) segmentation that reduces the computational costs by a factor of 25 with limited impact on the quality of results.
Semantic segmentation has a number of practical applications,
and for most such applications the computational costs are critical.
The method follows a typical two-column network structure,
where one column accepts an input image,
while the other accepts a half-resolution version of that image.
By identifying specific regions in the full-resolution image that can be safely ignored,
as well as carefully tailoring the network structure,
we can process approximately 15 high-resolution Cityscapes images~(1024$\times$2048) per second using a single GTX 980 video card,
while achieving a mean intersection-over-union score of 72.9\% on the Cityscapes test set.
\end{abstract}

\section{Introduction}
\begin{figure*}[t]
\begin{center}
\includegraphics[width=0.68595\linewidth,trim=0 130 285 0]{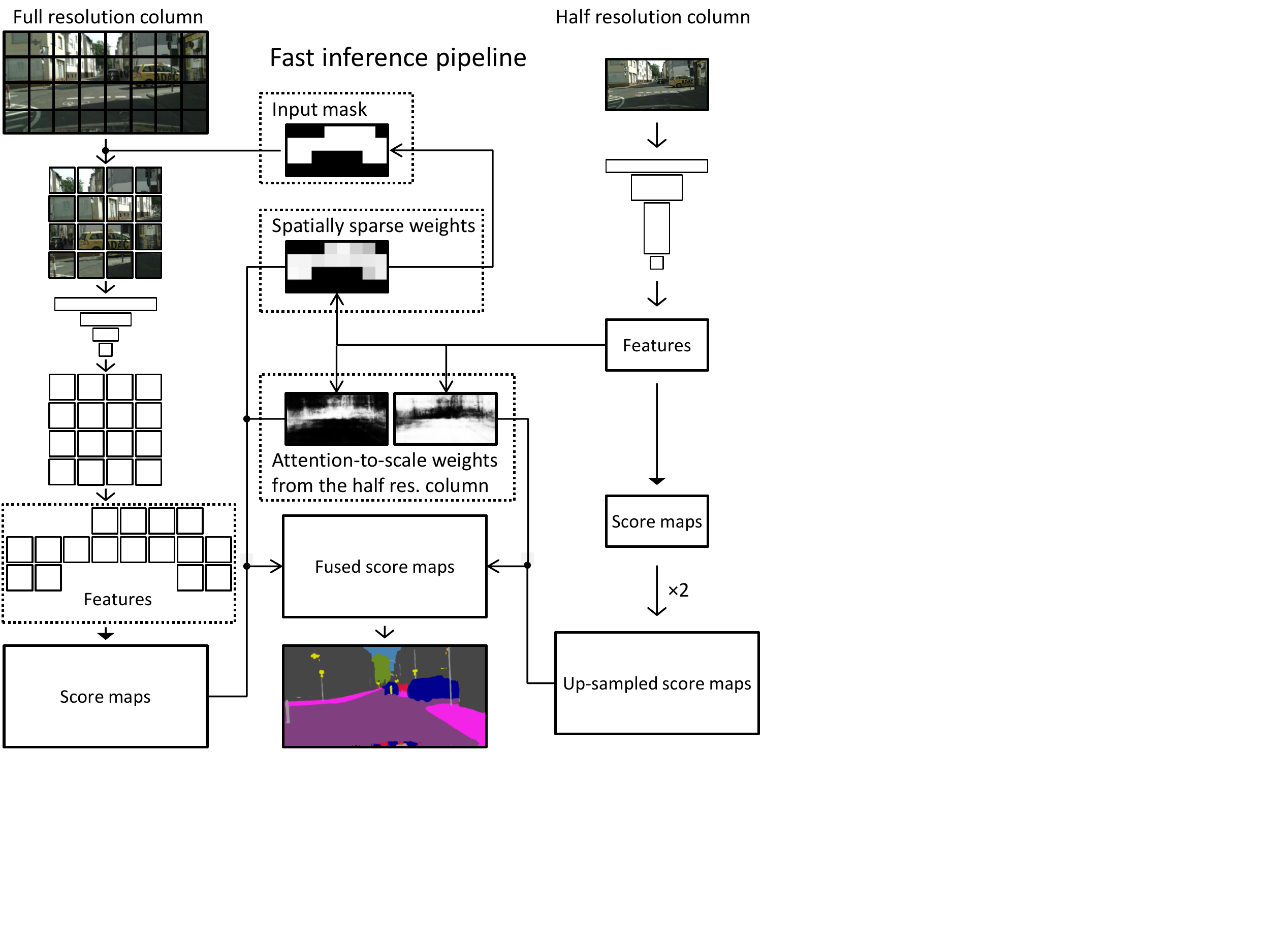}
\end{center}
\caption{
The inference pipeline of our proposed method.
Depending on the spatially sparse weights generated by the half-resolution column,
we skip part of the local regions in the full-resolution column.
}
\label{fig:inference}
\end{figure*}

Semantic (image) segmentation amounts to predicting the category of each pixel in a given image.
Take the task illustrated in Fig.~\ref{fig:inference} for example,
we are supposed to classify the pixels in an image recorded by a car-carried camera into 19 categories,
which include \textit{road}, \textit{car}, \textit{pedestrian}, \textit{bicycle}, etc.
Currently, most of the state-of-the-art approaches to this end are based on fully-convolutional networks (FCNs)~\cite{FCN.CVPR.2015.Long}.
A typical FCN-based method is usually equivalent to crop a local region for each pixel,
and then predict the category of that pixel using a fine-tuned backbone network.
However, FCNs can save computational costs largely by reusing the feature maps for adjacent image crops,
and so for their corresponding pixels.

Furthermore, two-column FCNs~\cite{Attention2Scale.2015.Chen} can capture the context of a pixel within local regions of different sizes.
Besides the regular full-resolution column, for example in Fig.~\ref{fig:inference},
the model has an additional column with a half-resolution input.
To make the final prediction,
we can calculate the element-wise sum/max of the score maps respectively generated from the two columns.
Or instead, we can first generate scale-aware weights and then calculate the weighted element-wise sum~\cite{Attention2Scale.2015.Chen}.

The observation that underpins this work is that most images contain large continuous regions belonging to the same category that can be segmented accurately at low resolutions.
Consider the two-column architecture in Fig.~\ref{fig:inference}.
The fact that the computational costs of the half-resolution column are only a quarter of those of the full-resolution column thus allows a significant improvement in efficiency.

The key to the model is a sparse weight map generated by the low-resolution column,
each entry of which corresponds to a square region in the original image.
Zero-valued entries in the low-resolution weight map indicate regions that do not need to be processed in high-resolution because the low-resolution classification will suffice.
Besides, the computational efficiency gained by not processing the indicated areas of the image at high-resolution are more than enough to compensate for the extra cost of generating the low-resolution weight map.

Although significant improvements in segmentation accuracy have been achieved recently~\cite{DeepLab2.2016.Chen,PSPNet.2016.Zhao},
improving the efficiency of FCNs has received far less attention.
This is despite the fact that it is one of the primary factors limiting their practical application.
Conventionally, datasets for general semantic segmentation have medium-sized images.
PASCAL VOC~\cite{PascalVoc.IJCV.2014.Everingham} for example, has images of no more than 500$\times$500 pixels.
The resolutions of typical cameras are far greater, but more importantly,
applications including driverless cars, and overhead image analysis demand a spatial acuity that a 500$\times$ 500 image cannot achieve.  

The Cityscapes~\cite{Cityscapes.CVPR.2016.Cordts} dataset is intended to provide images such as might be captured by a driverless car.
These images are of~1024$\times$2048 pixels, and are captured at approximately 15 frames per second.
The fact that many applications require high-resolution images to be captured and processed constantly drives significant practical demand for high pixel-throughput semantic segmentation.

In this work, our contributions are:
\begin{itemize}
\item
We introduce spatial sparsity into a typical two-column fully-convolutional network,
and show how to largely improve the inference speed without having a significant adverse impact on accuracy.
\item
We show that it is possible to reduce the computational cost of an indicative FCN by a factor of approximately 25,
at the cost of a small drop in accuracy, using spatial sparsity,
but also in-column and cross-column connections, and by removing residual units.
\end{itemize}

\section{Related work}
The proposed approach is closely related to two of the most active topics in computer vision,
i.e, residual networks~(ResNets) and semantic image segmentation using fully-convolutional networks (FCNs).

He~et~al.~\cite{ResNet.CVPR.2016.He,ResNet2.2016.He} proposed ResNets to combat the vanishing gradient problem during training very deep convolutional networks.
A ResNet is usually composed of multiple stacked residual units,
each of which either directly passes through or makes some modifications to its input.
ResNets have outperformed previous state-of-the-art models in various tasks, such as object detection~\cite{MNC.CVPR.2016.Dai} and semantic image segmentation~\cite{DeepLab2.2016.Chen}.
They have gradually replaced VGGNets~\cite{VGGNet.2014.Simonyan} in the computer vision community, as the standard feature extractors.
Nevertheless, note that the real mechanism underpinning the effectiveness of ResNets is not yet clear.
Veit~et~al.~\cite{UnraveledResNet.2016.Veit} claimed that they behave like ensembles of relatively shallow networks.
On the other hand, Greff~et~al.~\cite{UnrolledIterativeEstimation.ICLR.2017.Greff} interpreted a ResNet as learning unrolled iterative estimation.
They claimed that the first residual unit would learn a coarse estimation,
while the following units would learn to refine this estimation iteratively.

Most of the recent state-of-the-art approaches to semantic image segmentation are based on FCNs,
which were proposed by Long~et~al.~\cite{FCN.CVPR.2015.Long}.
FCNs soon became the mainstream approache to dense prediction tasks,
such as depth estimation and image super-resolution, primarily due to their efficiency.
On top of FCNs, Chen~et~al.~\cite{Attention2Scale.2015.Chen} fused the score maps obtained from two versions of an input image with different sizes using scale-aware weights.
Empirical results in the literature~\cite{DeepLab2.2016.Chen} also show that stronger pre-trained features~(using ResNets instead of VGGNets as the backbone networks) can further improve FCN performance.

\section{Method}
In this section we demonstrate the structure of our (single-column and two-column) baseline fully-convolutional networks (FCNs),
introduce spatial sparsity into the two-column version,
and finally describe the additional methods that we have adopted to further improve performance.

\subsection{Basic FCNs}
The low resolution of extracted feature maps is one of the challenges in applying traditional deep convolutional networks to semantic segmentation.
This problem originates from the need for a large field-of-view, while maintaining computational tractability, and results in a variety of down-sampling operations, including, for example, spatial pooling with a stride of two.

\begin{figure}[t]
\begin{center}
\includegraphics[width=0.90\linewidth,trim=0 260 300 0]{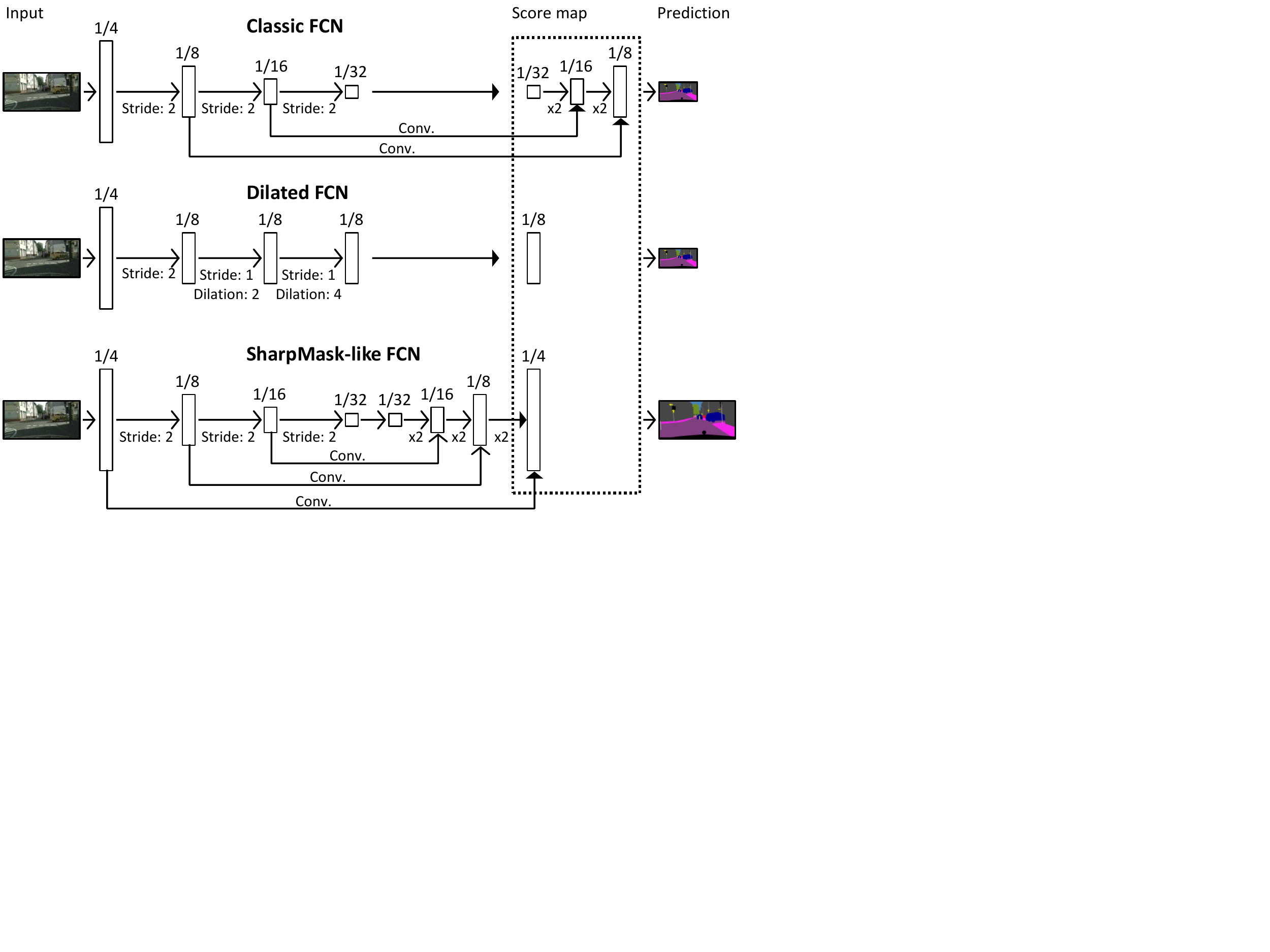}
\end{center}
\caption{
Three typical approaches to increasing the resolution of predicted score maps, i.e., classic fully-convolutional networks~\cite{FCN.CVPR.2015.Long}, dilated networks~\cite{DeepLab.ICLR.2015.Chen}, and our method derived from SharpMask~\cite{SharpMask.2016.Pinheiro}.
Note that the distinction between fusing the score maps directly and combining low-level and high-level features is the key difference between the classic approach and our SharpMask-like approach.
}
\label{fig:baseline}
\vspace{-3.0mm}
\end{figure}

We show three typical approaches to solving this problem in Fig.~\ref{fig:baseline}.
A classic FCN~\cite{FCN.CVPR.2015.Long} predicts multiple groups of score maps from feature maps of different resolutions,
and directly fuses (by summing) these score maps to make the final prediction.
While this works, it is not the most effective choice.
The DeepLab approaches~\cite{DeepLab.ICLR.2015.Chen,DeepLab2.2016.Chen} instead remove part of the down-sampling operations.
At the same time, they use dilated convolution kernels to ensure the consistency of field-of-view.
However, larger feature maps inevitably lead to increased computational costs,
which undermines our goal.
On the other hand, we in this work use a structure inspired by the SharpMask~\cite{SharpMask.2016.Pinheiro} model, which was originally proposed for instance segmentation.
Our empirical results show it offers a good compromise between efficiency and effectiveness.

\subsection{Two-column FCNs}
The size of the content to be recognized can vary significantly in semantic segmentation tasks,
which underpins the importance of multi-scale processing.
To this end, a typical pipeline includes the following three steps.
First, calculate the score maps for an input image;
second, down-sample that image and calculate another group of score maps;
and finally, combine the previous two groups of score maps.
Note that this can readily be generalized for cases with more than two groups of score maps.

\begin{figure}[t]
\begin{center}
\includegraphics[width=0.50\linewidth,trim=0 260 520 0]{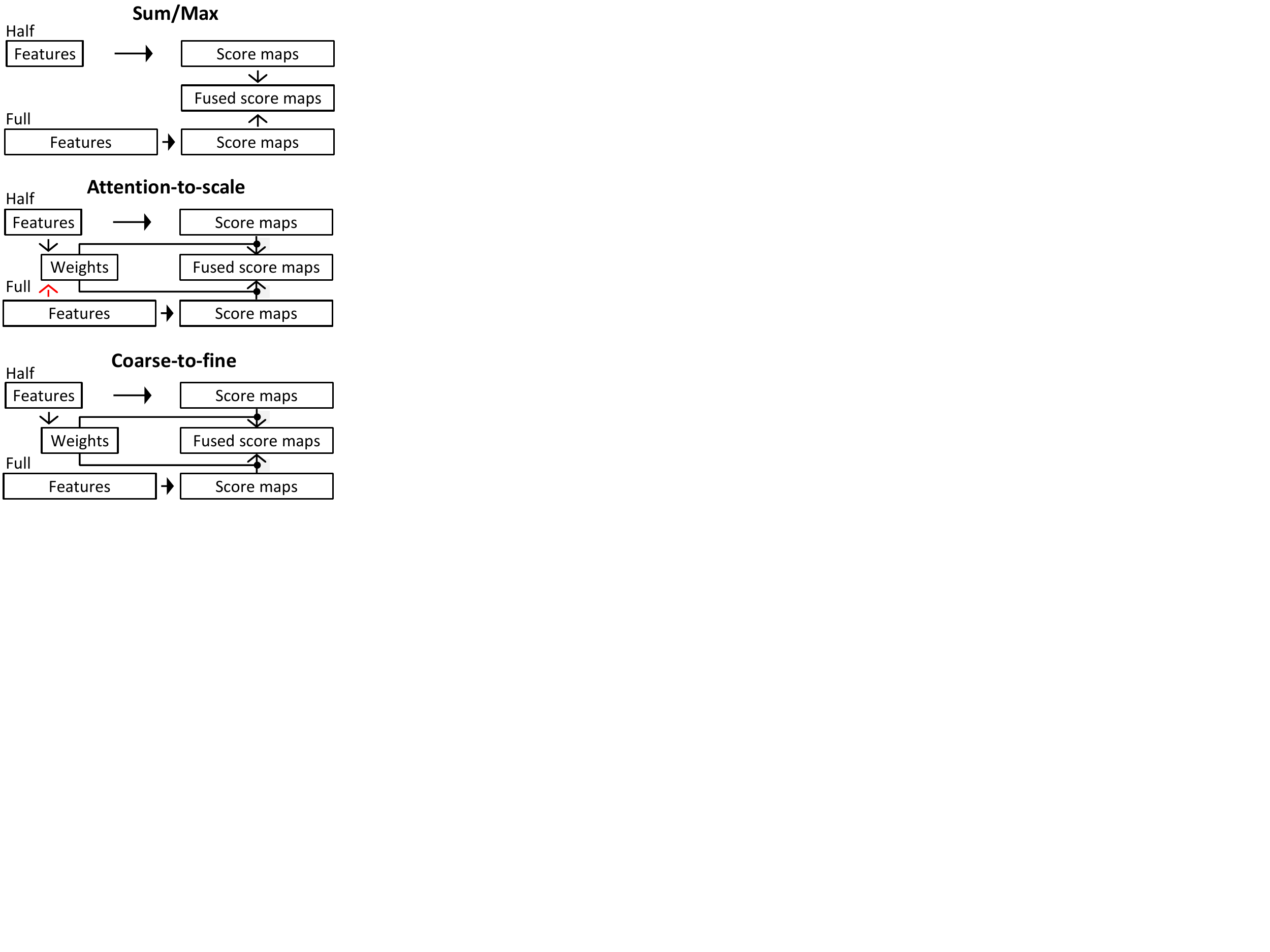}
\end{center}
\caption{
Four typical approaches to fusing the predicted score maps from two columns, i.e., sum/max~\cite{Attention2Scale.2015.Chen}, attention-to-scale~\cite{Attention2Scale.2015.Chen} and our coarse-to-fine.
Note that whether the scale-aware weights depend on the full-resolution column is the key difference between the classic attention-to-scale and our coarse-to-fine approaches.
}
\label{fig:baseline2}
\vspace{-3.0mm}
\end{figure}

We show four typical approaches to multi-scale fusion in Fig.~\ref{fig:baseline2}.
The most straightforward two choices are directly applying the element-wise summation or maximization to the score maps obtained from the two columns.
Another common approach in the literature is to first generate a group of (spatially varying) scale-aware weights,
then apply the Softmax activation function to ensure that the two entries at the same spatial location sum to one,
and finally calculate the weighted element-wise summation.
Chen~et~al.~\cite{Attention2Scale.2015.Chen} have evaluated all of the above three methods (represented by the first two diagrams in Fig.~\ref{fig:baseline2}).
They observed that the `attention-to-scale' approach performed the best in their proposed framework.
The key drawback of this approach,
particularly in the context of our model,
is the dependence of the scale-aware weights on the full-resolution column.
Given that full-resolution features do not exist for significant areas of the image,
they cannot inform the calculation of the scale-aware weights,
since much of the information required to achieve the red arrow in Fig.~\ref{fig:baseline2} is unavailable.
We thus adopt the illustrated `coarse-to-fine' approach instead,
whereby the scale-aware weights depend only on the half-resolution column.

\subsection{Spatial sparsity}
We here introduce sparsity to network activations but not weights.
One straightforward approach would be to apply a winner-take-all strategy,
which amounts to setting all activations to zero except for the largest value, or perhaps the $k$ largest values for some fixed $k$.
Another approach has been proposed in the context of restricted Boltzmann machines~\cite{TrainRBM.2010.Hinton},
which amounts to penalizing the cross entropy between the distribution of each activation (across different data points) and a specified binomial distribution.
Ideally, each activation will get activated (being one) with a specified probability $p$,
given different input data.
The first approach seems suboptimal but still works reasonably according to our experiments.
The second approach performs consistently better, however.

We can also encourage sparsity in the activations by sparse coding~\cite{SparseCodingNetwork.2017.Sun},
e.g., by penalizing their $\ell_1$ norm.
However, with this approach we cannot directly specify a target active probability.
The model may then compute zero or all of the local regions of an image in its full-resolution column.
This behaviour will lead to wide variations in inference times between different images,
which is a critical drawback in a real-time system.

We expect that some of the square regions in an image would be assigned with all-zero weights,
so we can skip these regions in the full-resolution column from the very beginning,
as illustrated in Fig.~\ref{fig:inference}.
Typically, given a local region in the original image,
there is a group of feature vectors extracted at its spatial location in the half-resolution column.
Suppose that an image can be evenly divided into $H \times W$ square regions,
and that there are $\tau^2$ vectors in the feature maps corresponding to each region.
We apply a single $\tau\times\tau$ convolution kernel with a stride of $\tau$ to those feature maps,
and obtain a weight matrix with $H \times W$ entries,
each of which corresponds to one square region in the original image.
Suppose that there are $N$ images per mini-batch,
then the obtained sparse weight $\boldsymbol{s}$ will have shape~$(N, H, W)$.
To ensure sparsity, we add a term~\cite{TrainRBM.2010.Hinton}
\begin{equation}
\lambda \cdot ( -p \log q - (1 - p) \log (1 - q) )
\end{equation}
to the overall loss function,
where $\lambda$ is a hyper-parameter scaling this term,
$p$ is the specified target probability,
and $q$ is a moving average probability.
This is computed as~\cite{TrainRBM.2010.Hinton}
\begin{equation}
q = \alpha \cdot q_{\textrm{old}} + (1 - \alpha) \cdot r,
\end{equation}
where the momentum $\alpha = 0.9$ (used throughout this work),
$q_{\textrm{old}}$ is the moving average probability used in the previous iteration,
and $r$ is the probability estimated only with the training data used in the current iteration.
However, instead of the convolution version derived from~\cite{TrainRBM.2010.Hinton}
\begin{equation}\label{eqn:origin_prob}
r_{y, x} = \frac{1}{N} \sum_{i=1}^{N} \operatorname{sigmoid}(s_{i, y, x}),
\end{equation}
we in this work use
\begin{equation}\label{eqn:new_prob}
r_{i} = \frac{1}{H \cdot W} \sum_{y=1}^{H} \sum_{x=1}^{W} \operatorname{sigmoid}(s_{i, y, x}).
\end{equation}

Note that the modification made here is critical,
because an image usually has a certain structure, especially  those in the Cityscapes~\cite{Cityscapes.CVPR.2016.Cordts} dataset.
There may be a large continuous region in the upper part (sky),
or another continuous region in the bottom part (road).
This observation suggests that the active probability of a region should vary as its spatial location changes.
Thus, in the context of this work,
instead of skiping the regions per location with a given probability, as in Eqn.~(\ref{eqn:origin_prob}),
it is more reasonable to skip the regions per image with a given probability, as in Eqn.~(\ref{eqn:new_prob}).

\begin{figure*}[t]
\begin{center}
\includegraphics[width=0.85\linewidth,trim=0 280 0 30]{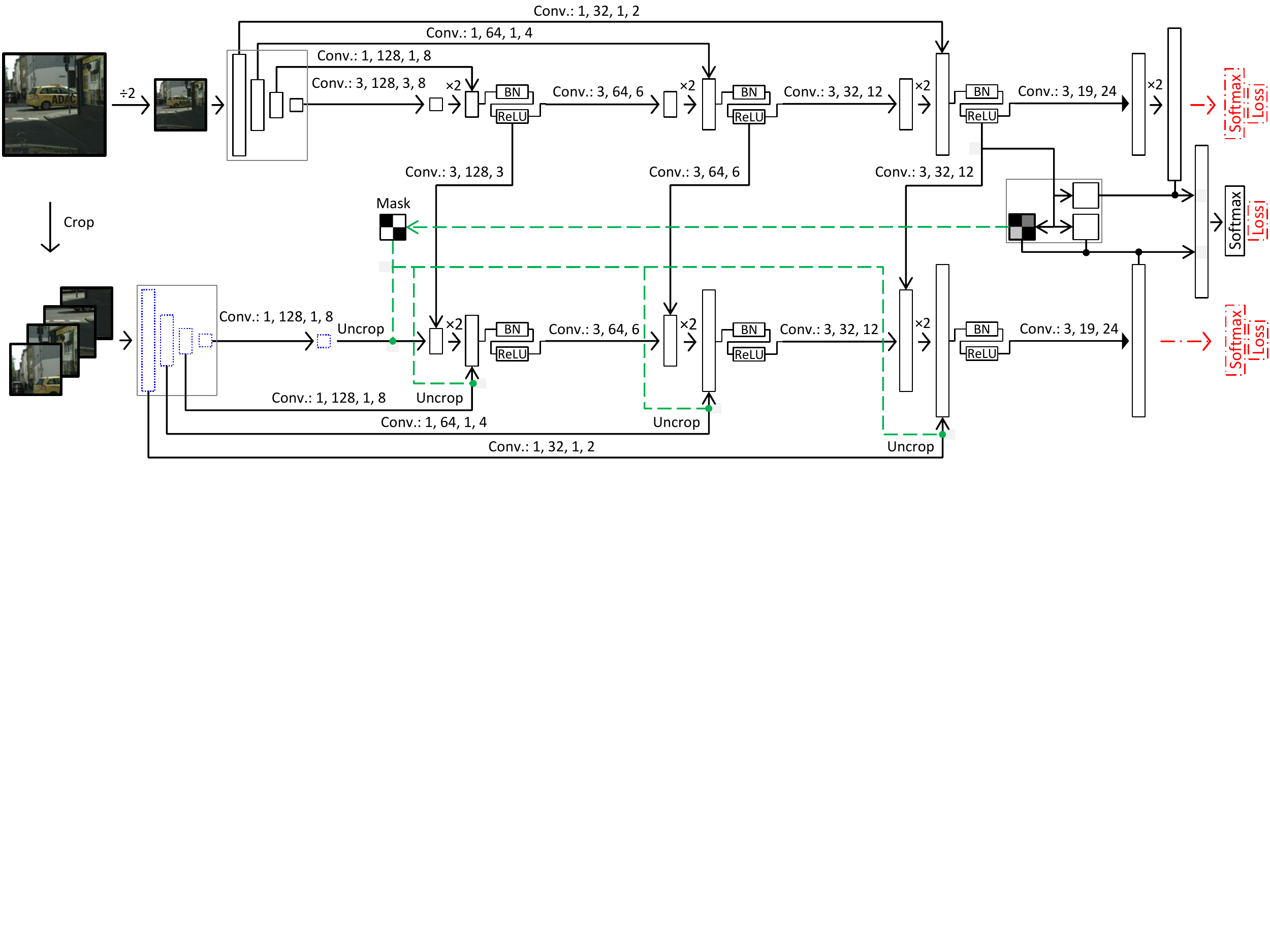}
\end{center}
\caption{
The structure of our model.
The configurations of layers are given as texts above the corresponding arrows.
For example, `Conv.: 3, 128, 3, 8' indicates a convolution layer having 128 kernels in the shape of 3$\times$3,
which are applied with a dilation rate of three and divided into eight groups~\cite{AlexNet.NIPS.2012.Krizhevsky}.
Blue dotted rectangles are feature maps of the cropped inputs;
red dash-dot arrows and rectangles denote the routes only enabled during training;
and back-propagation gradients never go through green dashed paths.
Best viewed in colour.
}
\label{fig:structure}
\vspace{-3.0mm}
\end{figure*}

\subsection{Network structure}\label{subsec:structure}
Our proposed model is an extension to the `coarse-to-fine' two-column structure illustrated in Fig.~\ref{fig:baseline2}, modified to achieve lower computational costs during inference.
Note that we name the pipeline in Fig.~\ref{fig:inference} `fast inference',
to distinguish it from the classic one.
The key distinction is that the fast-inference version only computes score maps for a part of an image in the full-resolution column.
To this end we, as usual, compute a pair of scale-aware weight maps,~i.e.,~$\boldsymbol{z}_1$ for the half and $\boldsymbol{z}_2$ for the full-resolution column.
We also compute an additional sparse weight map $\boldsymbol{s}$ following the steps described in the previous subsection.
We then up-sample $\boldsymbol{s}$ to $\boldsymbol{z}_2$'s shape,
and calculate the element-wise product of the two,
which is the sparse `coarse-to-fine' weight map required for the full-resolution column.
We also quantize the sparse weight map $\boldsymbol{s}$ into a mask,
which will be applied to the local regions cropped from the original image in the full-resolution column.
More details of our model are shown in Fig.~\ref{fig:structure}.
Note the two extra Softmax loss functions, as used by Chen~et~al.~\cite{Attention2Scale.2015.Chen},
and the in-column connections, as used in SharpMask~\cite{SharpMask.2016.Pinheiro}.
All the above components constitute our basic `sparse coarse-to-fine' model.

This basic model works well but loses too much accuracy when switching from classic to fast inference.
The cause of this problem is that an image is sliced into too small crops~(typically 256$\times$256),
so a lot of pixels~(near the boundary of an image crop) are classified without enough contexts during fast inference.
To avoid this problem, we make another two modifications to the basic model, as shown in Fig.~\ref{fig:structure}.
First, we also crop each input image during training.
This is undesirable for a plain two-column model,
due to the above demonstrated boundary problem.
However, it makes sense in our case here,
since we will crop the image anyway during fast inference.
In this way, we can at least extract consistent features in both the training and fast inference phases.
We process the obtained image crops separately as different samples,
and finally put their feature maps together to approximate those of the original image.
We call this operation `uncrop'.
This modification ensures a model to return exactly the same results either in the classic or fast inference phase,
together with the above described masking operation.
As illustrated in Fig.~\ref{fig:structure},
we `uncrop' the extracted features and in-column connections in the full-resolution column.
Second, we add several cross-column connections.
Since the half-resolution column covers a whole image,
reusing its features in the full-resolution column can help relieve the boundary problem,
We label this model as `improved sparse coarse-to-fine'.

We have also observed that the configuration of added layers can be optimized for better efficiency,
at the cost of a limited drop in accuracy,
as shown in Fig.~\ref{fig:structure}.
First, we gradually reduce the number of kernels as the feature maps grow spatially,
i.e., 128, 64 and 32.
Second, we introduce group sparsity~\cite{AlexNet.NIPS.2012.Krizhevsky},
i.e., eight groups for those layers directly on top of the raw features,
and respectively eight, four and two groups for those 1$\times$1 layers within the jump connections.
Third, we use a 1$\times$1 (but not 3$\times$3) layer on top of the raw features in the full-resolution column.

\subsection{Residual unit removal}
Removing part of the layers from top to bottom (those close to the output end) is the conventional way to reuse the features of a pre-trained network.
This method works well for classic models, e.g., AlexNets~\cite{AlexNet.NIPS.2012.Krizhevsky} and VGGNets~\cite{VGGNet.2014.Simonyan}, as well as ResNets~\cite{ResNet.CVPR.2016.He}.
Usually, we only remove the top-most layer, which is the linear classifier.
Furthermore, we can gradually reduce the computational costs by removing more and more layers, from top to bottom.
There is a trade-off, however, in that using fewer layers improves efficiency but removes high-level features.

Veit~et~al.~\cite{UnraveledResNet.2016.Veit} found that removing one arbitrary residual unit (with identity mappings) from a ResNet had little impact on its performance.
Besides, Greff~et~al.~\cite{UnrolledIterativeEstimation.ICLR.2017.Greff} claimed that a group of consecutive residual units can learn an unrolled iterative estimation,
supposing that all units (except for the first one) have identity mappings.
In other words, subsequent units are iteratively refining the coarse estimation learned by the first unit.
Based on these observations,
we can safely remove at most 12 of the 16 residual units in the ResNet-50 model~\cite{ResNet.CVPR.2016.He},
since there are only four residual units with non-identity mappings.
At the same time, the last of the four units outputs a coarse estimation of the high-level features learned by the original ResNet-50 model.
Our empirical results show that this technique steadily reduces the computational costs,
but sometimes has an adverse impact on the performance.
However, we find it possible to control this side effect using proper training strategies.

\begin{table*}[th]
\setlength{\tabcolsep}{5pt}
\begin{center}
\resizebox{0.85\textwidth}{!}{
\begin{tabular}{c|c|c|c|c|c|c|c|c|c|c|c}
\hline
\multicolumn{5}{c|}{Increasing dilations} & \multicolumn{3}{c|}{Up-sampling with jump connections} & \multirow{2}{*}{Pixel acc.} & \multirow{2}{*}{Mean acc.} & \multirow{2}{*}{Mean IoU} & \multirow{2}{*}{Calculations} \\
\cline{1-8}
1 & 2 & 3 & 4 & 5 & 1/32$\rightarrow$1/16 & 1/16$\rightarrow$1/8 & 1/8 $\rightarrow$ 1/4 & & & & \\
\hline\hline
& & & \checkmark & \checkmark & -- & -- & \checkmark & 95.67 & 82.61 & 74.83 & 786g \\
\hline
& & & & \checkmark & -- & \checkmark & \checkmark & 95.56 & 82.07 & 74.21 & 256g \\
\hline
& & & & & \checkmark & \checkmark & \checkmark & 95.40 & 81.64 & 73.78 & \textbf{165}g \\
\hline
\end{tabular}
}
\end{center}
\vspace{-2.0mm}
\caption{Results (\%) of networks with different configurations of kernel dilations and up-sampling layers on the Cityscapes~\cite{Cityscapes.CVPR.2016.Cordts} \textit{val} set.
All of the three models are SharpMask-like.
And structure of the third one is given in Fig.~\ref{fig:baseline},
which is chosen in this work for efficiency. 
}
\label{tbl:basic}
\vspace{-4.0mm}
\end{table*}

\section{Experiments}
We use the ResNet-50~\cite{ResNet.CVPR.2016.He} model to extract features throughout all the experiments.
Our implementation was carried out using the MXNet~\cite{MXNet.2015.Chen} tool-kit.
The actual inference speed, especially on GPU cards,
may vary largely depending on the implementation details.
We thus report the number of multiplication-adds per forward pass instead for the sake of fair comparison.

\subsection{Dataset}
We evaluate our method using the Cityscapes~\cite{Cityscapes.CVPR.2016.Cordts} dataset,
which consists of street scene photos taken by vehicle-carried cameras.
In total, there are 2,975 labelled images for training, another 500 for validation, i.e., the \textit{val} set.
There are also 1,525 testing images for which the ground-truth labels are kept secret to enable fair evaluation, i.e., the \textit{test} set.
There is also an additional set, consisting of 19,998 coarsely labelled images.
Pixels in an image may belong to 1 of 19 semantic categories, including \textit{road}, \textit{car}, \textit{pedestrian}, \textit{bicycle}, etc.
They further belong to seven super-categories, i.e., \textit{flat}, \textit{nature}, \textit{object}, \textit{sky}, \textit{construction}, \textit{human}, and \textit{vehicle}.
All images in this dataset are of the same size, which is very large, i.e., 1024$\times$2048.
On the \textit{val} set, we report,
1) the pixel accuracy, which is the percentage of correctly labelled pixels on a whole test set,
2) the mean pixel accuracy, which is the mean of class-wise pixel accuracies, and
3) the mean IoU score, which is the mean of class-wise intersection-over-union scores.
On the \textit{test} set, we only report the mean IoU scores.
\subsection{Basic FCNs}
Table~\ref{tbl:basic} shows results for three different basic FCNs.
The structure of the third model is illustrated in Fig.~\ref{fig:baseline},
labelled the `SharpMask-like FCN'.
We predict quarter resolution labels for the best accuracy, according to Wu~et~al.~\cite{InstanceSegmentation.2016.Wu}.
Namely, for a 1024$\times$2048 Cityscapes image,
we predict 19~(the number of valid categories) 256$\times$512 score maps,
and up-sample them back to the original resolution.
Our backbone network, i.e., ResNet-50~\cite{ResNet.CVPR.2016.He}, only calculates 1/32 resolution feature maps.
In the literature, there are two approaches to restoring the quarter resolution maps.
First, we can add up-sampling layers together with jump connections~\cite{FCN.CVPR.2015.Long} on top of the features.
Alternatively, we can remove the down-sampling operations and accordingly increase the kernel dilation rates in the following convolution layers, i.e., using the hole algorithm~\cite{DeepLab.ICLR.2015.Chen}.
Each approach has its advantages.
We can thus use both in the same network, as an intermediate choice.
We show how the prediction resolution is increased in the first two groups of columns in Table~\ref{tbl:basic}.
For example, in the first model, we increase the dilation rate by a factor of two both after the fourth and fifth down-sampling operations (both implemented by convolution layers) in ResNet-50;
and then we up-sample the obtained 1/8 resolution feature maps by a factor of two.
For all the three networks the added hidden feature layers consistently have 128 3$\times$3 kernels,
which are all applied with dilations so that the receptive fields have sizes around 200.
For example, the dilation rates for the three hidden feature layers in the third model are respectively 3, 6 and 12.
For the top-most linear classifier, the first network has 19 3$\times$3 kernels,
while the latter two only have 1$\times$1 kernels.

Table~\ref{tbl:basic} shows that to choose among these models is to compromise between effectiveness and efficiency.
Since we aim at a real-time model in this work,
the best trade-off is achieved by the third approach,~i.e.,
to use three up-sampling layers with jump connections.
We thus build our models using this approach for the following experiments.

\subsection{Impact of multi-scaling and fusing}\label{subsec:multi-column}

\begin{table*}[t]
\setlength{\tabcolsep}{3pt}
\begin{center}
\resizebox{0.85\textwidth}{!}{
\begin{tabular}{l|c|c|c|c|c|c|c|c}
\hline
\multirow{2}{*}{Fusion approach} & \multicolumn{4}{c|}{Input resolution} & \multirow{2}{*}{Pixel acc.} & \multirow{2}{*}{Mean acc.} & \multirow{2}{*}{Mean IoU} & \multirow{2}{*}{Calculations} \\
\cline{2-5}
& 1/4$\times$ & 1/2$\times$ & full & 2$\times$ & & & & \\
\hline\hline
\multirow{4}{*}{max} & & \checkmark & \checkmark & & 95.64 & 83.48 & 75.34 & 206g \\
\cline{2-9}
& \checkmark & \checkmark & \checkmark & & 95.62 & 82.57 & 74.30 & 216g \\
\cline{2-9}
& & & \checkmark & \checkmark & \textbf{95.68} & 82.84 & 75.16 & 824g \\
\cline{2-9}
& & \checkmark & \checkmark & \checkmark & 95.65 & 83.48 & 75.21 & 865g \\
\hline
sum & & \checkmark & \checkmark & & 95.59 & 82.69 & 75.20 & 206g \\
\hline
attention-to-scale & & \checkmark & \checkmark & & 95.64 & 82.62 & 75.22 & 206g \\
\hline
coarse-to-fine (CTF) & & \checkmark & \checkmark & & \textbf{95.68} & 82.94 & 75.14 & 206g \\
\hline
sparse coarse-to-fine (SCTF) & & \checkmark & \checkmark & & 95.43 & 82.37 & 74.72 & 206g \\
\hline
improved sparse coarse-to-fine (ISCTF) & & \checkmark & \checkmark & & 95.56 & \textbf{83.76} & \textbf{75.40} & 212g \\
\hline
\end{tabular}
}
\end{center}
\vspace{-2.0mm}
\caption{Results (\%) of networks with different input columns and fusion approaches on the Cityscapes~\cite{Cityscapes.CVPR.2016.Cordts} \textit{val} set.
Refer to Fig.~\ref{fig:baseline2} for structures of the first four fusion approaches;
and refer to Fig.~\ref{fig:structure} for the structures of the latter two.
This table shows that it is a good choice to only use half and full-resolution inputs,
and that all fusion approaches perform similarly.
Refer to Section~\ref{subsec:multi-column} for more details.
}
\label{tbl:multi-column}
\vspace{-3.0mm}
\end{table*}

We first study the impact of different input columns, with the `max' fusion method,
as shown in Table~\ref{tbl:multi-column}.
Doubling the input resolution almost quadruples the computational cost, from 206g to 824g,
but has no positive impact on the mean IoU score, which goes from 75.34\% to 75.16\%.
Adding another column, with a quarter or double resolution input,
does not improve the performance either.
Thus, in the following experiments, we use the two-column configuration with half and full-resolution inputs.
Although this increases the computational costs by about 25\%, from 165g to 206g,
we in return achieve better mean IoU scores.

We also find that the previously proposed fusion approaches,~i.e.,
`sum', `max' and `attention-to-scale', perform similarly,
which is in contrast to the observations of Chen~et~al.~\cite{Attention2Scale.2015.Chen}.
This difference probably stems from the different characteristics of the datasets,~i.e., Cityscapes~\cite{Cityscapes.CVPR.2016.Cordts}~vs.~PASCAL~VOC~2012~\cite{PascalVoc.IJCV.2014.Everingham},
and backbone networks,
i.e., ResNet-50~\cite{ResNet.CVPR.2016.He}~vs.~VGG-16~\cite{VGGNet.2014.Simonyan},
which is not our focus in this work.

Besides, our `coarse-to-fine'~(CTF) model also performs comparably to the above three approaches.
This result suggests that there is no need for the attention model to depend on the full-resolution column,
which is the key observation that enables us to speed-up the inference phase by introducing spatial sparsity.
In other words, we can skip some local regions of an input image in the full-resolution column to reduce the computational costs.
Introducing the spatial sparsity while still testing following the classic inference pipeline,
the `sparse coarse-to-fine'~(SCTF) approach performs slightly worse than CTF by 0.42\%.
However, the `improved sparse coarse-to-fine'~(ISCTF) approach outperforms CTF by 0.26\%.
This improvement probably originates from the cross-column connections,
which can enhance features in the full-resolution column using those features with relatively long-range context information obtained from the half-resolution column.

\subsection{Impact of spatially sparse testing}\label{subsec:sparse}
Table~\ref{tbl:sparse} shows that the proposed fast inference approach leads to a significant degradation in performance (about 3\%) for SCTF,
from 74.72\% to 71.74\%, due to the boundary problem.
However, ISCTF returns consistent results using either classic or fast inference.
Although we can specify a target active probability during training,
it still is a loose constraint which cannot fix the number of actually activated local regions.
Ideally, we should skip an arbitrary number of local regions per image in the full-resolution column,
depending on the number of zeros in its generated sparse weight map.
However, for implementation convenience,
we always skip the same number of regions per image,
by applying the winner-take-all strategy.
Skipping half of the local regions (16/32), we can consistently reduce the computational costs by about 35\% for SCTF, from 206g to 133g,
and about 34\% for ISCTF, from 212g to 139g.

The remaining notable points about the results in Table~\ref{tbl:sparse} are as follows.
First, the proposed method consistently outperforms the trivial `winner-take-all' strategy.
Second, $\lambda = 0.001$ seems the best setting here.
Third, although we can achieve better accuracy with $p = 0.5$, i.e., 75.40\% and 139g.
it also seems a good choice to let $p = 0.25$ for efficiency, i.e., 74.74\% and 103g,
which is the very setting we use in our following experiments.
Fourth, using too small local regions (128$\times$128) is harmful,
since it becomes harder to overcome the boundary problem.
\begin{table*}
\setlength{\tabcolsep}{2pt}
\begin{center}
\resizebox{0.85\textwidth}{!}{
\begin{tabular}{l|l|l|c|c|c|c|c|c|c}
\hline
\multicolumn{3}{l|}{Approach to sparsity} & Input size & Region size & Computed & Pixel acc. & Mean acc. & Mean IoU & Calc. \\
\hline\hline
\multicolumn{10}{c}{sparse coarse-to-fine (SCTF)} \\
\hline
sparse activations & $p=0.5$ & $\lambda=0.001$ & 1024$\times$2048 & 256 & 16/32 & 95.08 & 79.54 & 71.74 & 133g \\
\hline\hline
\multicolumn{10}{c}{improved sparse coarse-to-fine (ISCTF)} \\
\hline
\multirow{3}{*}{winner-take-all (WTA)} & $p=0.25$ & \multicolumn{1}{|c|}{--} & 1024$\times$2048 & 256 & 8/32 & 95.29 & 80.88 & 72.59 & 103g \\
\cline{2-10}
& $p=0.4$ & \multicolumn{1}{|c|}{--} & 1024$\times$2048 & 256 & 12/32 & 95.33 & 81.89 & 73.57 & 121g \\
\cline{2-10}
& $p=0.5$ & \multicolumn{1}{|c|}{--} & 1024$\times$2048 & 256 & 16/32 & 95.33 & 82.30 & 74.37 & 139g \\
\hline
\multirow{10}{*}{sparse activations} & $p=0.13$ & $\lambda=0.001$ & 1024$\times$2048 & 256 & 4/32 & 95.06 & 80.49 & 72.26 & 84.5g \\
\cline{2-10}
& $p=0.2$ & $\lambda=0.001$ & 1024$\times$2048 & 256 & 6/32 & 95.30 & 81.54 & 73.04 & 93.7g \\
\cline{2-10}
& $p=0.25$ & $\lambda=0.001$ & 1024$\times$2048 & 256 & 8/32 & 95.35 & 83.04 & \underline{\color{blue}74.74} & \underline{\color{blue}103}g \\
\cline{2-10}
& $p=0.4$ & $\lambda=0.001$ & 1024$\times$2048 & 256 & 12/32 & 95.42 & 83.49 & 75.16 & 121g \\
\cline{2-10}
& \multirow{4}{*}{$p=0.5$} & $\lambda=0.0001$ & 1024$\times$2048 & 256 & 16/32 & 95.53 & 83.33 & 74.97 & 139g \\
\cline{3-10}
&  & \multirow{2}{*}{$\lambda=0.001$} & 1024$\times$2048 & 128 & 64/128 & \textbf{95.60} & 83.37 & 75.02 & 139g \\
\cline{4-10}
&  &  & 1024$\times$2048 & 256 & 16/32 & 95.56 & \textbf{83.76} & \textbf{75.40} & 139g \\
\cline{3-10}
&  & $\lambda=0.01$ & 1024$\times$2048 & 256 & 16/32 & 95.54 & 83.55 & 75.19 & 139g \\
\hline
\end{tabular}
}
\end{center}
\vspace{-2.0mm}
\caption{Fast inference results (\%) of networks with different configurations of spatial sparsity on the Cityscapes~\cite{Cityscapes.CVPR.2016.Cordts} \textit{val} set.
Note the drastic drop in accuracy of SCTF compared to the one in Table~\ref{tbl:multi-column},
due to the boundary problem.
Besides, the trivial WTA method performs reasonably but is consistently inferior to our method with sparsity constraints to activations.
For efficiency, instead of the best performing configuration, i.e., $p=0.5$,
we next always use the $p=0.25$ setting.
Refer to Section~\ref{subsec:sparse} for more details.
}
\label{tbl:sparse}
\vspace{-4.0mm}
\end{table*}

\subsection{Impact of removing residual units}\label{subsec:removing}
We show the results of networks with different residual units removed in Table~\ref{tbl:removing}.
For efficiency, we choose relatively aggressive configurations here.
Notably, we can remove all units on the top-most level in the full-resolution column.
In this case, we also remove the following 1$\times$1 layer,
since there are no feature maps as its input any more.
However, this does not cause a fatal problem,
probably due to the first cross-column connection.
According to the results in Table~\ref{tbl:removing},
unit removal has an adverse impact on performance,
since it changes the inputs of many units.
However, the resulting gaps can be removed by online bootstrapping~\cite{InstanceSegmentation.2016.Wu} and adding in the Cityscapes~\cite{Cityscapes.CVPR.2016.Cordts} coarse set during training.
For those bad performing models with removed units,
we find that the generated attention weights for the half-resolution column sometimes dominate.
In this case, there is no contribution to the final prediction from the full-resolution column.
Probably, we remove units so aggressively in the full resolution column that this column can hardly converge to a reasonable point.
In this way, the attention model learns to discard this column and the whole model converges into a bad local minima.
However, fortunately our adopted strategies can help the model step out of this local minima.
Besides, as listed in the bottom-most row in Table~\ref{tbl:removing},
the model trained from scratch performs much worse than its pre-trained counterpart (63.54\% vs.~66.12\%),
which further justifies our method of unit removal.

\begin{table*}
\setlength{\tabcolsep}{3pt}
\begin{center}
\resizebox{0.90\textwidth}{!}{
\begin{tabular}{c|c|c|c|c|c|c|c|c|c|c|c|c|c|c|c|c|c}
\hline
\multicolumn{6}{c|}{1$\times$ column} & \multicolumn{6}{c|}{1/2$\times$ column} & \multicolumn{3}{c|}{ISCTF ($p=0.25$, $\lambda=0.001$)} & + online bs. & + coarse set & \multirow{2}{*}{Calc.} \\
\cline{1-17}
Removed & 1 & 2 & 3 & 4 & 5 & Removed & 1 & 2 & 3 & 4 & 5 & Pixel acc. & Mean acc. & Mean IoU & Mean IoU & Mean IoU & \\
\hline\hline
0 & -- & 3 & 4 & 6 & 3 & 0 & -- & 3 & 4 & 6 & 3 & 95.38 & 82.71 & 73.66 & 74.40 & 74.61 & 77.0g \\
\hline
13 & -- & 1 & 1 & 1 & 0 & 6 & -- & 2 & 3 & 4 & 1 & 94.89 & 78.21 & 69.93 & 71.30 & 74.09 & 33.8g \\
\hline
13 & -- & 1 & 1 & 1 & 0 & 6 & -- & 1 & 2 & 5 & 2 & 94.72 & 79.06 & 70.35 & 72.52 & 73.71 & 33.8g \\
\hline
12 & -- & 1 & 1 & 2 & 0 & 8 & -- & 1 & 2 & 4 & 1 & 94.70 & 77.90 & 69.73 & 70.65 & 74.31 & 31.5g \\
\hline
12 & -- & 1 & 1 & 1 & 1 & 8 & -- & 1 & 2 & 4 & 1 & 94.68 & 77.72 & 69.35 & -- & -- & 31.3g \\
\hline
13 & -- & 1 & 1 & 1 & 0 & 8 & -- & 1 & 2 & 4 & 1 & 94.73 & 77.82 & 69.63 & 71.34 & -- & 29.3g \\
\hline
13 & -- & 1 & 1 & 1 & 0 & 10 & -- & 1 & 1 & 3 & 1 & 94.34 & 74.59 & 66.12 & 68.53 & 73.20 & 24.7g \\
\hline\hline
-- & -- & 1 & 1 & 1 & 0 & -- & -- & 1 & 1 & 3 & 1 & 93.87 & 72.63 & 63.54 & 65.33 & 70.73 & 24.7g \\
\hline
\end{tabular}
}
\end{center}
\caption{Results (\%) of networks with different configurations of removed residual units on the Cityscapes~\cite{Cityscapes.CVPR.2016.Cordts} \textit{val} set.
All models have the optimized structure shown in Fig.~\ref{fig:structure}.
The bottom-most model is trained from scratch.
Note the marginal gains of the top-most model~(with no unit removed) achieved by online bootstrapping~\cite{InstanceSegmentation.2016.Wu} and enlarging training data,
compared to those for the remaining ones.
This shows that the two adopted strategies can effectively remove the side effect of the unit removal.
Refer to Section~\ref{subsec:removing} for more details.
}
\label{tbl:removing}
\vspace{-1.0mm}
\end{table*}

\subsection{Ablation study}
We show the respective costs and contributions of the components in Table~\ref{tbl:ablation}.
On top of the ISCTF model,
spatial sparsity together with fast inference reduces 51.4\% of the computations costs,
but only with a 0.6\% loss of accuracy.
And network structure optimization further reduces the costs by 25.2\%.
On top of the ISCTF model trained with online bootstrapping~\cite{InstanceSegmentation.2016.Wu} and enlarged training data,
our Model A reduces the computational costs by 59.0\%, with a 1.03\% loss of accuracy.
This cost is approximately 25 times less than that of the baseline model, i.e., 31.5g~v.s.~786g.
\begin{table*}
\setlength{\tabcolsep}{5pt}
\begin{center}
\resizebox{0.90\textwidth}{!}{
\begin{tabular}{l|l|c|c|c|c|c|c|r|r}
\hline
\multicolumn{2}{l|}{Approach} & Pixel acc. & Mean acc. & Mean IoU & $\Delta$ & Overall $\Delta$ & Calc. & \multicolumn{1}{c|}{$\Delta$} & \multicolumn{1}{c}{Overall $\Delta$} \\
\hline\hline
\multicolumn{2}{l|}{single-column model} & 95.67 & 82.61 & 74.83 & -- & -- & 786g & -- & -- \\
\hline\hline
\multicolumn{10}{c}{two-column models} \\
\hline
\multicolumn{2}{l|}{max} & 95.64 & 83.48 & 75.34 & -- & -- & 206g & -- & -- \\
\hline
\multicolumn{2}{l|}{SCTF ($p=0.5$)} & 95.43 & 82.37 & 74.72 & $-0.62$ & $-0.62$ & 206g & -- & -- \\
\hline
\multicolumn{2}{l|}{ISCTF ($p=0.25$)} & 95.35 & 83.04 & 74.74 & $+0.02$ & $-0.60$ & 212g & $+2.9\%$ & $+2.9\%$ \\
\cline{3-10}
\multicolumn{1}{l}{\phantom{12}} & + fast inference & -- & -- & -- & -- & -- & 103g & $-51.4\%$ & $-50.0\%$ \\
\cline{3-10}
\multicolumn{1}{l}{\phantom{12}} & + structure optimizing & 95.38 & 82.71 & 73.66 & $-1.08$ & $-1.68$ & 77.0g & $-25.2\%$ & $-62.6\%$ \\
\cline{3-10}
\multicolumn{1}{l}{\phantom{12}} & + online bootstrapping & 95.51 & 81.93 & 74.40 & $+0.74$ & $-0.94$ & -- & -- & -- \\
\cline{3-10}
\multicolumn{1}{l}{\phantom{12}} & + Cityscapes coarse set & 95.53 & 83.50 & 74.61 & $+0.21$ & $-0.73$ & -- & -- & -- \\
\cline{2-10}
\multicolumn{1}{l}{\phantom{12}} & Model A: 1-1-2-0-1-2-4-1 & 95.46 & 82.91 & 74.31 & $-0.30$ & $-1.03$ & 31.5g & $-59.0\%$ & $-84.7\%$ \\
\hline
\end{tabular}
}
\end{center}
\caption{Results (\%) of ablation study and comparison with our implemented baseline single-column model (doubling the dilation factor twice and up-sampling with jump connections only once as listed in Table~\ref{tbl:basic}) on the Cityscapes~\cite{Cityscapes.CVPR.2016.Cordts} \textit{val} set.
}
\label{tbl:ablation}
\vspace{-1.0mm}
\end{table*}

\subsection{Comparison with previous methods}
We compare with previous methods in the literature,
especially those with inference speed reported,
as listed in Table~\ref{tbl:comparison}.
Here, our reported speed is obtained using the built-in profiler of MXNet~\cite{MXNet.2015.Chen},
evaluated using a PC having one i7-4790 CPU and one Geforce GTX 980 video card installed with CUDA 8.0 and cuDNN 5.1.
Among the (almost) real-time approaches,
our method outperforms the previous best one with a clear margin, i.e., 14.6\%.
There is a concurrent work to ours released via arXiv (labelled ICNet by Zhao~et~al.~\cite{ICNet.2017.Zhao}),
which is similar to our model in the sense that it also follows a multi-column structure.
However, we here rely on spatial sparsity and the new style of removing units,
which were not present in their work.
Besides, they also applied model compression, which may be a complementary technique to our method.

\begin{table}
\begin{center}
\resizebox{0.40\textwidth}{!}{
\begin{tabular}{l|c|c|c}
\hline
Method & mean IoU & Time & Speed (fps) \\
\hline\hline
FCN-8s~\cite{FCN.CVPR.2015.Long} & 65.3\% & 500ms & 2.0 \\
DeepLab-v2~\cite{DeepLab2.2016.Chen} & 71.4\% & 625ms & 1.6 \\
RefineNet~\cite{RefineNet.2016.Lin} & 73.6\% & -- & $\sim$0.9 \\
PSPNet~\cite{PSPNet.2016.Zhao} & 80.2\% & -- & $\sim$0.5 \\
\hline
SegNet~\cite{SegNet.2015.Badrinarayanan} & 57.0\% & 60ms & 16.7 \\
ENet~\cite{ENet.2016.Paszke} & 58.3\% & 13ms & 76.9 \\
\hline
Ours & 72.9\% & 68ms & 14.7 \\
\hline
\end{tabular}
}
\end{center}
\caption{Comparison with previous best performers on the Cityscapes~\cite{Cityscapes.CVPR.2016.Cordts} \textit{test} set,
especially those with inference speeds reported.
Note that the speeds are only for reference,
since the implementation details vary significantly between different works.
}
\label{tbl:comparison}
\vspace{-3.0mm}
\end{table}

\subsection{Qualitative results}
\begin{figure}[t]
\begin{center}
\includegraphics[width=0.65\linewidth,trim=50 340 420 25]{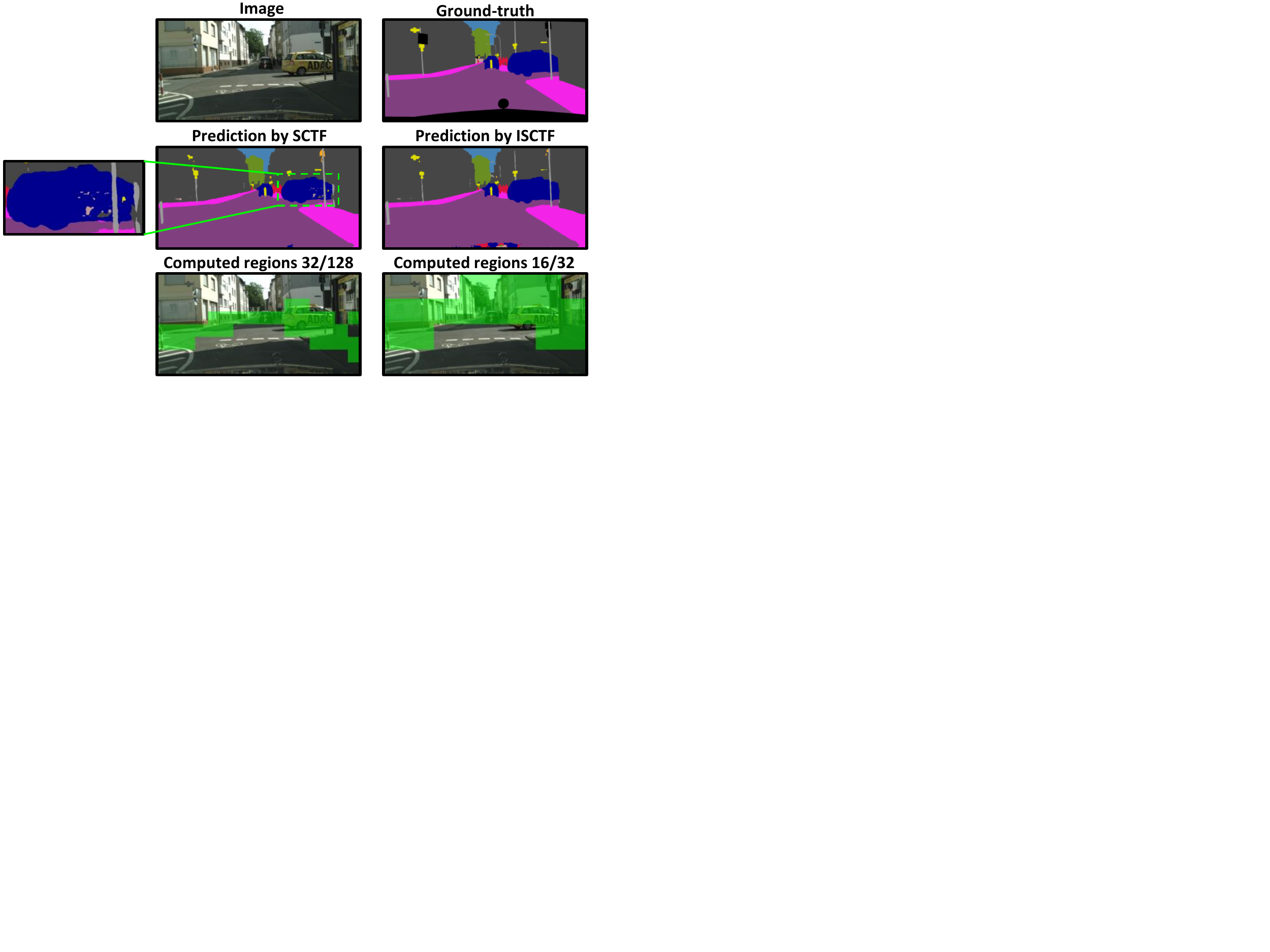}
\end{center}
\caption{
Qualitative results.
Second row: Comparison between SCTF and ISCTF, showing the boundary problem due to insufficient contexts.
Third row: Actually computed local regions in the full-resolution column.
Better viewed in colour.
}
\label{fig:qualitative}
\vspace{-4.0mm}
\end{figure}

We show qualitative results in Fig.~\ref{fig:qualitative}.
An example of the boundary problem discussed in Section~\ref{subsec:structure} is shown in the second row.
Due to lack of context information, the SCTF model failed to identify some pixels inside the yellow car,
while the ISCTF model is able to figure it out.
We also show two examples of the activated local regions in the full-resolution column in the third row.
These are reasonably activated regions,
since there usually are small pedestrians or vehicles around them,
which require being processed in high resolutions.
To skip as many regions as possible (for efficiency), the left case may be preferable.
Since the regions have smaller sizes,
we can more accurately locate the hard regions.
However, on the other hand, using smaller regions may aggravate the boundary problem.

\section{Conclusion}
We in this work have proposed a real-time and accurate approach to semantic image segmentation.
To this end, we have introduced spatial sparsity into a carefully-tailored two-column network,
and studied the impact of removing residual units to the performance of our method.
Finally, we have achieved a 25 times reduction of computational costs, only for a 2.0\% loss of accuracy on the Cityscapes \textit{val} set.

{\small
\bibliographystyle{ieee}
\bibliography{ref}
}

\end{document}